\documentclass[conference]{IEEEtran}
\IEEEoverridecommandlockouts

\usepackage{cite}
\usepackage{amsmath,amssymb,amsfonts}
\usepackage{graphicx}
\usepackage{booktabs}
\usepackage{multirow}
\usepackage{array}
\usepackage[table]{xcolor}
\usepackage{tikz}
\usetikzlibrary{arrows.meta,positioning,fit,backgrounds,calc}
\usepackage{url}
\usepackage[hidelinks]{hyperref}

\definecolor{grpSota}{HTML}{E7F0FA}   % comparison vs prior  淡蓝
\definecolor{grpAbl}{HTML}{E9F5E9}    % all other groups     淡绿
\definecolor{oursrow}{HTML}{ECECEC}   % our main model 行    浅灰
\definecolor{bestclr}{HTML}{1A4FA0}   % best value 深蓝
\newcommand{\best}[1]{\textbf{\textcolor{bestclr}{#1}}}  % best per column
\definecolor{caseHdr}{HTML}{2C3E50}   % 两个表统一的深色表头底
\definecolor{caseAlt}{HTML}{EEF3F8}   % cases 斑马行 淡蓝灰
\newcommand{\ev}[2]{\leavevmode\rlap{\footnotesize\textbf{#1.}}\hspace{1.3em}#2\par}

\begin{document}

\title{DeliMF: Gated Cross-Attention Fusion of Structured Time Series and Clinical Text for Early Intensive Care Delirium Prediction}

\author{
  Yibo Yang$^{1\dagger}$ \quad
  Zerui Huang$^{1\dagger}$ \quad
  Haomiao Yang$^{2}$ \quad
  Zixia Wang$^{3*}$ \\[4pt]
  $^{1}$Independent Researcher, China \\
  $^{2}$Xinjin District People's Hospital, China \quad
  $^{3}$University of Exeter, United Kingdom \\[4pt]
  {\small \texttt{yang.374960202@gmail.com} \quad \texttt{huangruirui92@gmail.com}} \\
  {\small \texttt{cdhmyang@126.com} \quad \texttt{zw483@exeter.ac.uk}} \\[6pt]
  {\small $^{\dagger}$These authors contributed equally to this work and share first authorship.} \\
  {\small $^{*}$Corresponding author.}
}
\maketitle

\begin{abstract}
Delirium affects a large fraction of intensive care unit (ICU) patients and foreshadows longer stays, lasting cognitive decline, and higher mortality. Flagging the patients who will become delirious within the first day of admission would give clinicians a window to intervene. Yet state-of-the-art predictors rely only on the structured electronic health record (EHR) of vital signs, laboratory values, and medications, while overlooking the free-text clinical notes written in the same window, which carry information the numbers cannot encode. To this end, we present DeliMF, a multimodal framework that fuses the structured time series with clinical text. It encodes the structured signals with a temporal Transformer and the time-stamped radiology reports with a clinical text encoder, and fuses the two through a gated cross-attention module. The gate learns a per-patient weight on the text; when a patient has no note the gate closes and DeliMF reduces to its structured-only form, giving robustness to missing text by construction. On an incident-only MIMIC-IV cohort, DeliMF improves significantly over the baselines, particularly on patients who carry clinical text, while losing nothing on those without. Ablation studies confirm the contribution of multimodal fusion and gated cross-attention, and a case study further shows that fusing clinical text with the structured stream recovers delirium cases that the structured data alone misses. We hope DeliMF encourages early ICU prediction to make fuller use of the overlooked clinical text and contributes to the early identification and timely intervention of delirium.
\end{abstract}


%==============================================================
\section{Introduction}
%==============================================================
% Delirium is an acute, fluctuating disturbance of attention and awareness, and it
% ranks among the most common neuropsychiatric complications of critical illness.
% It is far from benign. Critically ill survivors who experience delirium carry a
% higher burden of long-term cognitive impairment, on the order seen after moderate
% traumatic brain injury~\cite{pandharipande2013brainicu}; in older patients more
% broadly, delirium predicts prolonged hospitalization and
% death~\cite{inouye2006delirium}. The case for \emph{early} prediction follows
% directly, because many precipitants are modifiable. Sedation can be lightened,
% sleep protected, and deliriogenic drugs withdrawn, but only for patients
% identified before the episode rather than after. A model that ranks risk within
% the first 24 hours of ICU admission is therefore the actionable target, and it is
% the target we adopt.
Delirium is an acute and fluctuating disturbance of attention and awareness, and it ranks among the most common neuropsychiatric complications of critical
illness~\cite{stollings2021delirium,wilson2020delirium,kotfis2023intensive}. Delirium in critically ill survivors is linked to long-term cognitive impairment comparable to moderate traumatic brain injury~\cite{pandharipande2013brainicu}, and in older patients more broadly it predicts higher mortality~\cite{inouye2006delirium}. Many precipitants of delirium are modifiable. Intervening before onset, for instance by protecting sleep and withdrawing deliriogenic drugs, can reduce its incidence and shorten its
course~\cite{sosnowski2023effect}. Such interventions are only meaningful before an episode begins. Flagging high-risk patients before onset is therefore of considerable value, since early risk identification can guide the timely use of
preventive measures~\cite{wassenaar2015predeliric}.

Delirium risk modeling has progressed from hand-crafted clinical scores~\cite{van2012development} to
machine learning. E-PRE-DELIRIC~\cite{wassenaar2015predeliric} estimates risk
from a handful of admission variables and remains a common clinical reference.
PRIDE~\cite{hur2021pride} and the dynamic model~\cite{gong2023predicting} push
performance on structured features with gradient boosting and neural networks
respectively. DeLLiriuM~\cite{contreras2025dellirium} reframes the task for a
large language model by serializing the electronic health record (EHR) into a
prompt~\cite{yang2026aidabench}. All of these methods build on structured EHR, the numerical record of vitals, labs, and medications. They provide effective tools for ICU delirium prediction.

However, clinical text carries information that is equally valuable~\cite{van2009risk} yet left unused by prior methods. In other ICU prediction tasks, integrating multimodal clinical data has been shown to improve performance~\cite{lyu2023multimodal,deznabi2021predicting}. 
The notes written in the first hours of an ICU stay
routinely name the very factors that drive delirium but are not recorded in the EHR, for example: a head CT that reads ``subarachnoid hemorrhage,'' a history of ``alcoholic cirrhosis,'' or a radiograph showing ``acute pneumonia.'' We aim to exploit this overlooked information, together with the structured data, for delirium prediction.

\begin{figure*}[!t]
\centering
\includegraphics[width=\textwidth]{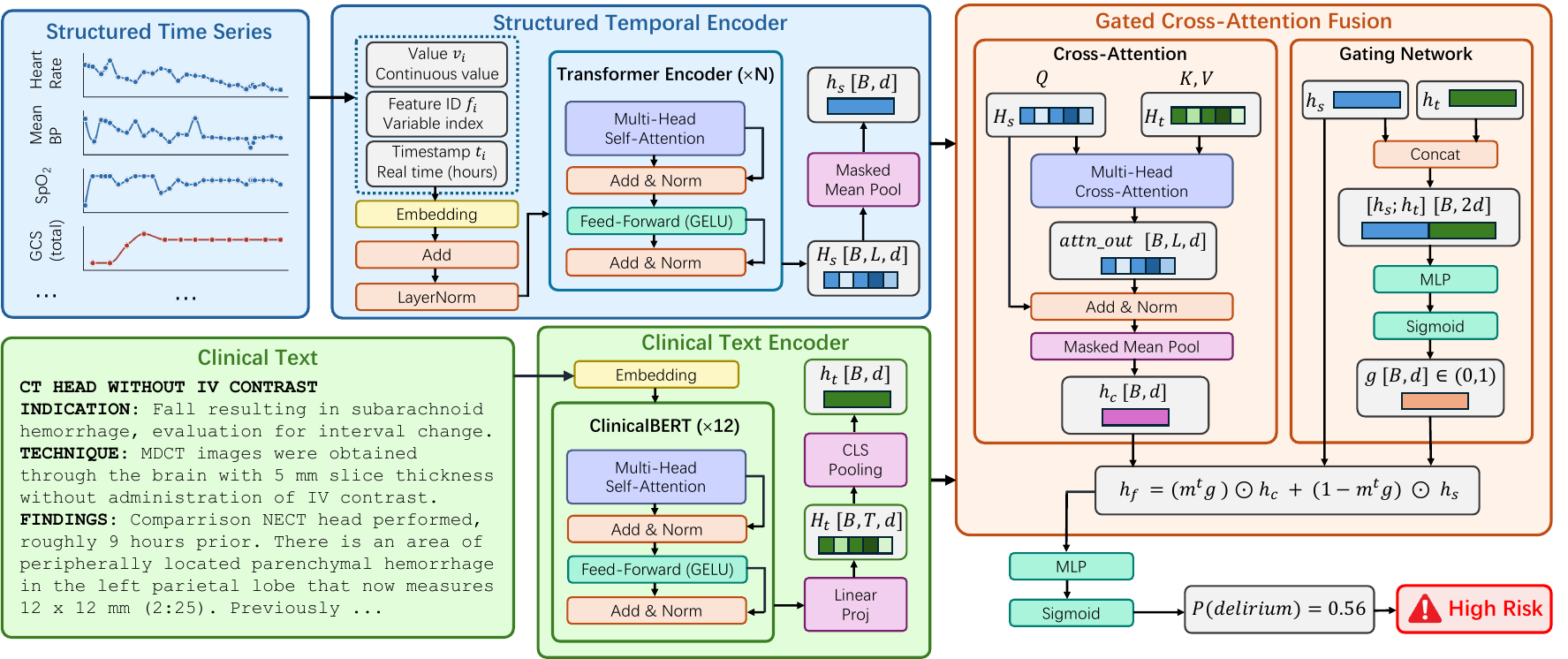}
\caption{Overview of \textbf{DeliMF}, traced through a real MIMIC-IV case
(61-year-old admitted after a fall, incident-delirium label). The structured
tower embeds each event from its value, feature id, and timestamp and encodes
the sequence with a temporal Transformer, pooled to $h_s$. The text tower encodes
the first-24\,h radiology report with ClinicalBERT, pooled to $h_t$. In the gated
cross-attention module, the structured states query the text, and a per-patient
gate weights the two representations into the fused $h_f$; when no note is
available the gate closes and DeliMF reduces to its structured-only form. A two-layer MLP
head maps $h_f$ to the delirium probability. For this case the report names a
subarachnoid hemorrhage with diffuse axonal injury, precipitants absent from the
tabular record, and DeliMF raises the probability to 0.56 while the
structured-only baseline assigns 0.21 and misses the case.}
\label{fig:overview}
\end{figure*}

Building on this observation, we propose \textbf{DeliMF} (\textbf{Deli}rium
\textbf{M}ultimodal \textbf{F}usion), a new framework that fuses two modalities,
the structured time series and the clinical text.
Figure~\ref{fig:overview} traces our full pipeline through a representative case.
Specifically, DeliMF has three parts. It models the structured stream with a
temporal Transformer that leverages the actual timestamp of each observation
rather than aligning the data to a fixed temporal grid. For the text stream, it
uses only radiology reports recorded prior to the prediction time, so as to
avoid information that would be unavailable at that time. The two representations
are then fused by a gated cross-attention module, in which the structured
representation queries the text for descriptive context and a gate weights the
textual and structured representations per patient.

Finally, we conduct extensive experiments to validate and evaluate DeliMF. On a cohort restricted to incident cases in the ICU, DeliMF improves significantly
over the baselines. We further conduct ablation studies that isolate the
contribution of multimodal fusion and gated cross-attention, analyze the learned
per-patient gate, and present a case study tracing the true delirium cases that
DeliMF recovers over the structured-only model, showing that clinical text indeed
provides effective information for delirium prediction beyond the structured data.

% Figure 1 框架图（内联，原 fig_overview.tex）

Overall, our contributions are summarized as follows:
\begin{itemize}
\item We identify and exploit the clinical text overlooked by prior prediction. Early clinical text records many precipitants that drive delirium yet resist structured encoding, and we introduce this text as a separate modality for prediction.
\item We propose \textbf{DeliMF}, a multimodal fusion framework based on gated
cross-attention. It models the structured stream as a native time series and the
text with a clinical text encoder, and weights the two representations through a
per-patient gate.
\item We conduct extensive experiments, including comparison with baselines,
ablation studies, and a case study, that validate DeliMF and show that fusing
clinical text with the structured stream through gated cross-attention yields
gains beyond either modality alone and beyond naive fusion.
\end{itemize}

%==============================================================
\section{Related Work}

\textbf{ICU delirium prediction.} Existing ICU delirium prediction methods are
built mostly on structured EHR. The E-PRE-DELIRIC
model~\cite{wassenaar2015predeliric} predicts delirium risk at admission with a
compact set of clinical variables, prioritizing interpretability and bedside use.
Later work replaces this fixed scoring with learned predictors.
PRIDE~\cite{hur2021pride} develops machine-learning models over EHR-derived features,
and a dynamic neural model~\cite{gong2023predicting} tracks risk through the stay.
DeLLiriuM~\cite{contreras2025dellirium} takes a different route, casting the
structured record into a textual prompt so that a large language model can score
it. A separate line of work models the structured stream as a native irregularly
sampled time series, encoding the real time between observations rather than a
fixed grid~\cite{shukla2019interpolation,tipirneni2022self}. In contrast, DeliMF reads the clinical text alongside the structured
stream and combines the two modalities through gated fusion.

\textbf{Multimodal clinical learning.} Multimodal fusion is an active direction in deep
learning~\cite{lou2026multitasks,lou2024mr,jin2026hypase} and is well developed for
modalities such as image and text, typically aligning and fusing the
representations of different modalities through contrastive learning or
cross-modal attention~\cite{radford2021learning,li2021align}.
In the clinical domain, this idea has been used to bring in clinical text beyond
the structured EHR. Some work~\cite{deznabi2021predicting,lyu2023multimodal,
shukla2020integrating} jointly models physiological time series and clinical notes
for in-hospital mortality and readmission, generally finding that text supplies
context the structured fields cannot encode. A complementary line learns to
weight each modality's contribution with a gate rather than fixing the fusion
scheme~\cite{arevalo2017gated}, and gating has also been applied to irregular
multimodal EHR~\cite{zhang2022improving}, though there the gate resolves
irregularity within each modality while the cross-modal fusion itself rests on
attention. DeliMF differs from these works in
two respects. The first is the clinical setting, as it targets early, incident
ICU delirium prediction. The second is the model structure, as it models the
structured stream as a native time series and places a per-patient gate directly
across the fused text and structured representations, so the gate both adapts
modality reliance to each patient and closes to recover the structured-only model
when no note is available.

%==============================================================
%==============================================================
% Method section lives in method.tex (edit that file, not here).
%==============================================================
% Method section — DeliMF
% This file is \input{} into main.tex. Edit ONLY this file for Method changes.
% All macros/colors/packages are defined in main.tex's preamble.
%==============================================================
\section{Method}
%==============================================================
\subsection{Problem formulation}
For each ICU stay we observe the first 24 hours after admission and predict
whether the patient develops delirium \emph{after} that window. The cohort is
\emph{incident-only}: stays already delirious or comatose within the first 24
hours are excluded, so the task is genuine prediction rather than detection. Let
a stay be represented by a structured event sequence $X^{s}$ and, when
available, a clinical text $X^{t}$; the label $y\in\{0,1\}$ marks subsequent
incident delirium. DeliMF estimates $P(y=1\mid X^{s}, X^{t})$.

\subsection{Structured temporal encoder}
The structured stream $X^{s}$ is a sequence of $L$ irregularly timed events, where each
event $i$ is a triple $(v_i, f_i, t_i)$ of a continuous value
$v_i\in\mathbb{R}$, a feature index $f_i\in\{1,\dots,F\}$ identifying one of the
$F$ structured variables, and a timestamp $t_i\in\mathbb{R}$. In the spirit of
Transformer input construction, each event is mapped to a token embedding
$x_i\in\mathbb{R}^{d}$ as the sum of three terms:
\begin{equation}
x_i \;=\; v_i\,w_{f_i} \;+\; E_{f_i} \;+\; \mathrm{TE}(t_i),
\label{eq:struct-token}
\end{equation}
i.e.\ a value term, a feature term, and a time term, where
$E\in\mathbb{R}^{F\times d}$ is a learned per-feature embedding table
indexed by $f_i$, $w_{f_i}\in\mathbb{R}^{d}$ is the per-feature value projection
selected by the same index, and the timestamp is encoded by a learnable
continuous function
\begin{equation}
\mathrm{TE}(t) \;=\; \mathrm{MLP}\bigl([\sin(\omega t);\cos(\omega t)]\bigr),
\label{eq:time-enc}
\end{equation}
with $\omega$ a vector of learnable frequencies; the sine and cosine terms are
concatenated and mapped to the $d$-dimensional token space by the MLP, so that the
encoding
reflects the
real spacing between observations rather than a fixed grid. Each stay is padded
to a common length $L$ with a binary mask $m\in\{0,1\}^{B\times L}$ marking real
events; stacking the token embeddings of $X^{s}$ over a batch of $B$ stays
gives $X\in\mathbb{R}^{B\times L\times d}$, which an $N$-layer
Transformer encoder~\cite{vaswani2017attention} processes under this padding mask:
\begin{equation}
\begin{aligned}
H_{s} &\;=\; \mathrm{TransformerEnc}(X)\in\mathbb{R}^{B\times L\times d},\\
h_{s,b} &\;=\; \frac{\sum_{\ell=1}^{L} m_{b,\ell}\,H_{s,b,\ell}}{\sum_{\ell=1}^{L} m_{b,\ell}}\;\in\;\mathbb{R}^{d},
\end{aligned}
\end{equation}
i.e.\ token states $H_{s}$ and, for each stay $b$, a mean over its valid
(non-padded) tokens; stacking the $h_{s,b}$ gives the pooled summary
$h_{s}\in\mathbb{R}^{B\times d}$.

\subsection{Clinical text encoder}
The radiology reports written within the first 24 hours are ordered by time and
concatenated into a single text $X^{t}$, encoded with a clinical text encoder; we
use ClinicalBERT~\cite{alsentzer2019clinicalbert} as the default and study a
Qwen2.5~\cite{qwen2025qwen25} alternative in Section~\ref{sec:backbone}. Writing
$Z=\mathrm{Encoder}(X^{t})$ for the contextual states and $\mathrm{Proj}$ for a
linear map to the shared dimension $d$, the token states are the projected
sequence and the pooled vector is the projected summary state:
\begin{equation}
\begin{aligned}
H_{t} &\;=\; \mathrm{Proj}(Z)\in\mathbb{R}^{B\times T\times d},\\
h_{t}^{(b)} &\;=\; \mathrm{Proj}\bigl(z_{b}\bigr)\in\mathbb{R}^{d},\\
z_{b} &\;=\;
\begin{cases}
Z_{b,1}, & \text{ClinicalBERT (\texttt{[CLS]})},\\
Z_{b,\tau_b}, & \text{Qwen (last token }\tau_b\text{)},
\end{cases}
\end{aligned}
\end{equation}
where $\tau_b$ is the last non-padded position of stay $b$. Stacking the
$h_{t}^{(b)}$ gives $h_{t}\in\mathbb{R}^{B\times d}$. The two backbones differ
in how this summary is read: the encoder-only ClinicalBERT uses its
\texttt{[CLS]} state, whereas the decoder-only Qwen uses the last token, whose
causal receptive field spans the whole report.

\subsection{Gated cross-attention fusion}
The two streams meet in the fusion module. Structured token states query the
text states through multi-head cross-attention, letting physiological trajectory
attend to descriptive context. The structured states form the queries and the
text states form the keys and values. Each of the $n_h$ heads projects them with
learned matrices $W_j^{Q},W_j^{K},W_j^{V}\in\mathbb{R}^{d\times d_k}$ into a
$d_k=d/n_h$ dimensional subspace,
\begin{equation}
Q_j=H_{s}W_j^{Q},\qquad K_j=H_{t}W_j^{K},\qquad V_j=H_{t}W_j^{V},
\end{equation}
and attends over the text with a scaled dot-product, masking padded and absent
text tokens:
\begin{equation}
\mathrm{head}_j \;=\; \mathrm{softmax}\!\Bigl(\frac{Q_j K_j^{\top}}{\sqrt{d_k}}+M\Bigr)V_j,
\label{eq:cross-head}
\end{equation}
where $M\in\{0,-\infty\}^{B\times L\times T}$ is an additive mask, shared across
heads, that suppresses padded or absent text tokens before the softmax,
\begin{equation}
M_{b,\ell,t}=
\begin{cases}
0, & \text{text token $t$ of stay $b$ is valid},\\
-\infty, & \text{otherwise},
\end{cases}
\end{equation}
so that a masked key receives zero attention weight. The heads
are concatenated and projected back to the shared dimension by an output matrix
$W^{O}\in\mathbb{R}^{d\times d}$,
\begin{equation}
\tilde{H}_{c} \;=\; \bigl[\mathrm{head}_1;\dots;\mathrm{head}_{n_h}\bigr]W^{O}
\in\mathbb{R}^{B\times L\times d},
\label{eq:cross-concat}
\end{equation}
and stabilized by a residual connection and layer normalization,
\begin{equation}
H_{c} \;=\; \mathrm{LayerNorm}\bigl(\tilde{H}_{c} + H_{s}\bigr),
\end{equation}
followed by masked mean pooling to $h_{c}\in\mathbb{R}^{B\times d}$. Rather than
mixing the two streams with a fixed coefficient, a gate conditioned on both
pooled summaries decides per patient how much of the cross-attended evidence to
admit. The gate is a two-layer MLP on the concatenated summaries, and the
text-availability flag $m^{t}_{b}\in\{0,1\}$ (one if stay $b$ has a note) modulates
the fused representation:
\begin{equation}
\begin{aligned}
g   &\;=\; \sigma\bigl(W_2\,\mathrm{ReLU}(W_1\,[\,h_{s};\,h_{t}\,] + b_1) + b_2\bigr)\in(0,1)^{d},\\
h_{f} &\;=\; (m^{t}g)\odot h_{c} + \bigl(1-m^{t}g\bigr)\odot h_{s}.
\end{aligned}
\label{eq:gate}
\end{equation}
Because $g$ is a nonlinear function of the inputs $[h_{s};h_{t}]$ and acts per
dimension, it is an input-adaptive, per-patient weight on the text-informed
representation rather than a shared constant. The second-layer bias $b_2$ is
initialized negative ($b_2\!=\!-2$, giving $g\!\approx\!\sigma(-2)\!=\!0.12$ at the
start of training), so the model leans on the structured stream first and admits
text only as it proves useful. When a stay has no note, $m^{t}_{b}\!=\!0$ zeroes
the gated term, giving $h_{f}\!=\!h_{s}$ exactly, so DeliMF reduces precisely to
its structured-only form. This gives missing-modality robustness by
construction. A two-layer MLP head maps $h_{f}$ to a delirium probability
$\hat{y}=\sigma(\mathrm{MLP}(h_f))\in[0,1]$, and the model is trained by a
class-weighted binary cross-entropy that counteracts the low positive prevalence:
\begin{equation}
\mathcal{L} \;=\; -\,\frac{1}{B}\sum_{b=1}^{B}
\bigl[\,w_{+}\,y_b\log\hat{y}_b + (1-y_b)\log(1-\hat{y}_b)\,\bigr],
\label{eq:loss}
\end{equation}
where the sum runs over the $B$ stays in a batch and the positive weight
$w_{+}$ is the ratio of negative to positive stays in the training set.

\subsection{Two-stage training}
Trained end-to-end from scratch, the high-capacity text encoder destabilizes
optimization and contaminates even the temporal pathway. We therefore train in two
stages: \emph{(i)} freeze the text encoder and train the structured encoder,
fusion, and head; \emph{(ii)} unfreeze the upper text-encoder layers and fine-tune
jointly at a smaller learning rate. Combined with the negative gate bias, this
lets the temporal pathway settle before the text is introduced as an increment.
Section~\ref{sec:comparison} reports the effect of this schedule against an
end-to-end variant.

%==============================================================
% Experiments section — DeliMF
% This file is \input{} into main.tex. Edit ONLY this file for Experiments changes.
% All macros/colors (est, \prob, v, colors) are defined in main.tex preamble.
%==============================================================
\section{Experiments}
%==============================================================
\subsection{Data and cohort}
\begin{figure}[t]
\centering
\includegraphics[width=\columnwidth]{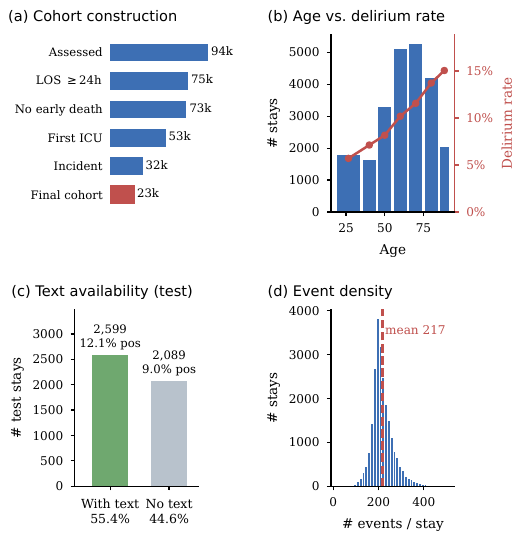}
\caption{Data and cohort overview. (a) Cohort construction from all
brain-status-assessed ICU stays down to the incident-only cohort of 23{,}431; the
final (red) bar is the analyzed cohort. (b) Age distribution (bars) and delirium
rate (line), which rises monotonically with age. (c) Text availability on the
test set: 55.4\% of stays carry a radiology report, and the with-text subgroup has
a higher positive rate. (d) Distribution of structured events per stay
(mean 217).}
\label{fig:cohort}
\end{figure}

We build the cohort from MIMIC-IV~\cite{johnson2023mimiciv} together with its note
module. Delirium labels are reconstructed from charted
CAM-ICU~\cite{ely2001camicu} features under the BRAIN-ICU three-state scheme of
normal, delirium, and coma~\cite{pandharipande2013brainicu}. To avoid label
contamination, an ``unable to assess'' reading is treated as coma rather than
negative. The reconstructed labels agree with the native delirium-assessment item
on $93.8\%$ of scored intervals, confirming their reliability. We then restrict
the cohort to incident-only stays, excluding those with prevalent delirium or coma
in the first 24 hours, early death, or a non-first ICU admission. The resulting
cohort comprises $23{,}431$ stays at a $10.7\%$ positive prevalence and is split by
patient into training, validation, and test sets, as summarized in
Fig.~\ref{fig:cohort}a.

The test set contains $4{,}688$ stays. Of these, $2{,}599$ stays, or $55.4\%$,
carry a radiology report time-stamped within the prediction window, while the
remaining $44.6\%$ have no accompanying text (Fig.~\ref{fig:cohort}c). Because
text is restricted to notes whose time stamp precedes the prediction point, the
setup is leakage-safe. Two further properties of the cohort are worth noting.
Delirium incidence rises monotonically with age, from $6.4\%$ below age 35 to
$15.1\%$ above age 85 (Fig.~\ref{fig:cohort}b), and each stay contributes $217$
structured events on average (Fig.~\ref{fig:cohort}d).

\subsection{Implementation}
The structured encoder uses a model dimension of $d{=}256$ with 8 attention heads
and 4 layers. We optimize with AdamW under a cosine schedule with warmup and a
class-weighted loss, and apply differentiated learning rates: $10^{-4}$ for the
structured and fusion parameters and $2\times10^{-5}$ for the text encoder. Every
model is run with three random seeds, and we report the mean and standard
deviation across seeds. Statistical significance against the baselines is assessed
with a paired bootstrap of $n{=}1000$ resamples, drawing test cases with
replacement.

\subsection{Metrics}
We report two threshold-free areas,
\begin{equation}
\begin{aligned}
\mathrm{AUROC} &\;=\; \int_{0}^{1}\!\mathrm{TPR}\;d(\mathrm{FPR}),\\
\mathrm{AUPRC} &\;=\; \int_{0}^{1}\!\mathrm{Prec}\;d(\mathrm{Rec}).
\end{aligned}
\label{eq:auroc-auprc}
\end{equation}
AUROC~\cite{hanley1982roc} remains stable under class imbalance, whereas
AUPRC responds to the positive prevalence~\cite{davis2006prroc}. On our skewed
cohort, whose positive rate is only $10.7\%$, the precision-recall view is
therefore the more informative one~\cite{saito2015prc}. We accordingly treat
\textbf{AUPRC} as the primary metric and read AUROC as a secondary,
optimism-prone signal. At a $0.5$ threshold we additionally report sensitivity,
specificity, and F1, and we summarize calibration with the Brier
score~\cite{brier1950score}, for which lower values are better.

\subsection{Comparison with prior delirium-prediction methods}
\label{sec:comparison}
Table~\ref{tab:main} compares DeliMF against the structured-only model and two
published baselines reimplemented under our cohort and protocol: the
DeLLiriuM-style serialization model~\cite{contreras2025dellirium} and an
E-PRE-DELIRIC-style predictor~\cite{wassenaar2015predeliric}. On AUROC the full
model and the structured-only model are statistically indistinguishable. On AUPRC,
which is the metric that matters under heavy imbalance, DeliMF pulls clearly ahead.
The contrast with the DeLLiriuM-style baseline is sharper still. That model
serializes the structured record into a prompt and discards its temporal order,
and DeliMF improves AUROC over it by more than eight points at $p\approx0$. We read
this as direct evidence that the structured stream is best served by native
temporal modeling rather than by language-model framing. This comparison also puts
the published DeLLiriuM result in perspective. Its reported external AUROC of
$0.824$~\cite{contreras2025dellirium} was obtained on a different cohort under a
different label definition and
is thus not directly comparable to ours; once cohort and protocol are held fixed,
the serialization approach falls behind. Text alone is a weak predictor, which is
unsurprising, as radiology reports are an indirect signal and cover only part of
the cohort. The text modality earns its place only when fused with the structured
stream, not when substituted for it.

% 主对比表（Overall，全部模型，6 指标）。内联，原 results_main_table.tex。
% GRU row: only AUROC/AUPRC reported in text (0.790/0.349); other cols n/a.
\begin{table*}[t]
\caption{Comparison on the MIMIC-IV test set (3 seeds, mean\,$\pm$\,std; threshold 0.5).
The table is organized into blocks: comparison with prior methods, single-modality
ablation, fusion design, training schedule, and encoder substitution. Our full model
is shaded; best per column in \best{bold blue} (lower is better for Brier).}
\label{tab:main}
\centering\footnotesize
\setlength{\tabcolsep}{5pt}
\renewcommand{\arraystretch}{1.18}
\begin{tabular}{l cccccc}
\toprule
\rowcolor{caseHdr}
\textcolor{white}{\textbf{Model}} & \textcolor{white}{\textbf{AUROC}} & \textcolor{white}{\textbf{AUPRC}} & \textcolor{white}{\textbf{F1}}
& \textcolor{white}{\textbf{Sens.}} & \textcolor{white}{\textbf{Spec.}} & \textcolor{white}{\textbf{Brier}} \\
\midrule
\rowcolor{grpSota}
\multicolumn{7}{@{\hspace{3pt}}l}{\emph{Comparison with prior methods}} \\
\midrule
\rowcolor{oursrow}
DeliMF (ours)          & \best{0.803}\,$\pm$\,0.003 & 0.360\,$\pm$\,0.012 & \best{0.391}\,$\pm$\,0.018 & 0.718\,$\pm$\,0.036 & 0.762\,$\pm$\,0.037 & 0.162\,$\pm$\,0.016 \\
DeLLiriuM-style        & 0.721\,$\pm$\,0.000 & 0.242\,$\pm$\,0.001 & 0.316\,$\pm$\,0.006 & 0.545\,$\pm$\,0.037 & 0.770\,$\pm$\,0.032 & 0.182\,$\pm$\,0.015 \\
E-PRE-DELIRIC-style    & 0.692\,$\pm$\,0.000 & 0.189\,$\pm$\,0.000 & 0.283\,$\pm$\,0.000 & 0.624\,$\pm$\,0.000 & 0.666\,$\pm$\,0.000 & 0.228\,$\pm$\,0.000 \\
\midrule
\rowcolor{grpAbl}
\multicolumn{7}{@{\hspace{3pt}}l}{\emph{Single-modality ablation}} \\
\midrule
Structured-only        & 0.800\,$\pm$\,0.002 & 0.334\,$\pm$\,0.007 & 0.390\,$\pm$\,0.003 & 0.663\,$\pm$\,0.057 & 0.791\,$\pm$\,0.031 & \best{0.148}\,$\pm$\,0.013 \\
Text-only              & 0.602\,$\pm$\,0.022 & 0.169\,$\pm$\,0.012 & 0.246\,$\pm$\,0.016 & 0.327\,$\pm$\,0.077 & \best{0.843}\,$\pm$\,0.047 & 0.223\,$\pm$\,0.030 \\
\midrule
\rowcolor{grpAbl}
\multicolumn{7}{@{\hspace{3pt}}l}{\emph{Fusion design}} \\
\midrule
Concat fusion          & 0.784\,$\pm$\,0.006 & 0.348\,$\pm$\,0.019 & 0.377\,$\pm$\,0.022 & 0.657\,$\pm$\,0.078 & 0.775\,$\pm$\,0.068 & 0.166\,$\pm$\,0.030 \\
Late fusion            & 0.782\,$\pm$\,0.003 & 0.331\,$\pm$\,0.013 & 0.371\,$\pm$\,0.005 & 0.636\,$\pm$\,0.036 & 0.784\,$\pm$\,0.024 & 0.165\,$\pm$\,0.013 \\
\midrule
\rowcolor{grpAbl}
\multicolumn{7}{@{\hspace{3pt}}l}{\emph{Training schedule} (two-stage is our default, top row)} \\
\midrule
DeliMF (end-to-end)    & 0.795\,$\pm$\,0.004 & 0.359\,$\pm$\,0.005 & 0.368\,$\pm$\,0.036 & 0.699\,$\pm$\,0.083 & 0.736\,$\pm$\,0.096 & 0.183\,$\pm$\,0.054 \\
\midrule
\rowcolor{grpAbl}
\multicolumn{7}{@{\hspace{3pt}}l}{\emph{Encoder substitution}} \\
\midrule
Qwen2.5 (text encoder)       & 0.801\,$\pm$\,0.004 & \best{0.361}\,$\pm$\,0.004 & 0.374\,$\pm$\,0.023 & \best{0.722}\,$\pm$\,0.066 & 0.738\,$\pm$\,0.067 & 0.176\,$\pm$\,0.028 \\
GRU (struct encoder)         & 0.790 & 0.349 & -- & -- & -- & -- \\
\bottomrule
\end{tabular}
\end{table*}

\subsection{DeliMF design ablation}
\label{sec:ablation-design}
DeliMF rests on three design choices: multimodal fusion, token-level
cross-attention, and the input-adaptive gate. We isolate the first two by
replacing each with a simpler alternative while holding every other component
fixed, using three seeds and the two-stage schedule throughout. The gate is
examined separately, through its missing-modality behavior in
Section~\ref{sec:backbone} and its learned distribution in
Section~\ref{sec:interp}. All results in this analysis are reported in
Table~\ref{tab:main}.

\noindent\emph{Multimodality.} Removing the text branch reduces DeliMF to its
structured-only form and gives back the AUPRC it had gained, a loss of roughly two
and a half points at $p\approx0.06$. Removing the structured branch instead
leaves the text-only model, whose AUROC collapses by some twenty points. The two
modalities are therefore not interchangeable: text complements structure but
cannot stand in for it.

\noindent\emph{Fusion design.} Replacing gated cross-attention with simple
concatenation costs close to two AUROC points at $p<0.001$, and late fusion fares
no better. This exposes a real risk rather than a marginal preference. When fused
naively, a weak modality can drag performance below the structured-only model, and
avoiding exactly this failure is what the gate is designed for.

\noindent\emph{Cross-attention.} Holding the fusion structure fixed while feeding
it the pooled text vector in place of token-level cross-attention costs about two
AUPRC points. The fine-grained interaction between physiological trajectory and
descriptive context thus contributes measurably and is not a decorative part of
the design.

Taken together, each design choice is justified by these results, as removing any
one of them brings DeliMF down to, or below, the structured-only AUPRC.

\subsection{Backbone sensitivity}
\label{sec:backbone}
Beyond DeliMF's design choices, we verify that the gains are not artifacts of any
single backbone. Replacing the temporal Transformer with a bidirectional
GRU~\cite{cho2014gru} costs a little over one AUROC point and leaves AUPRC
essentially unchanged. This gap is small and non-significant, yet it still
supports our emphasis on native temporal modeling. The text encoder behaves
similarly. Substituting ClinicalBERT with Qwen2.5-1.5B~\cite{qwen2025qwen25}
leaves whole-cohort performance
unchanged and nudges the with-text AUROC slightly upward, indicating that a
stronger pretrained language model yields a text representation at least as useful
for this task.

\subsection{Subset analysis}
\begin{figure}[t]
\centering
\includegraphics[width=\columnwidth]{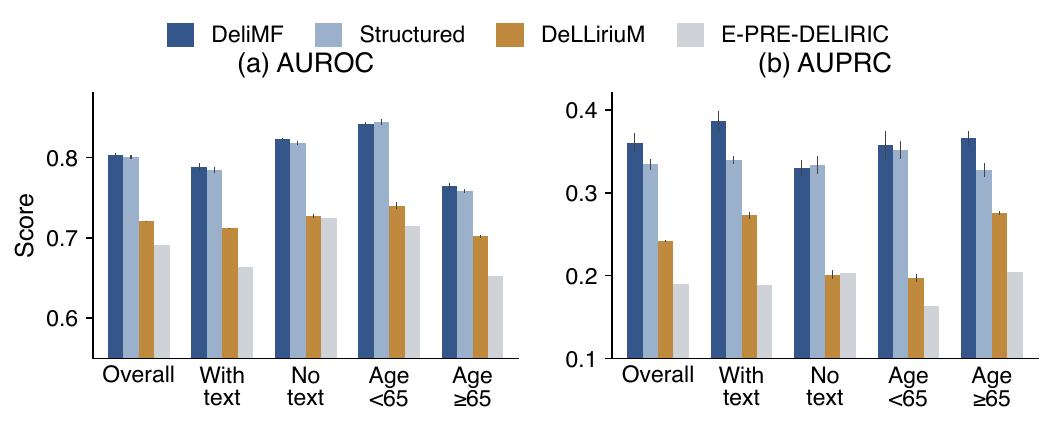}
\caption{Subset analysis across text availability and age (3 seeds,
mean$\pm$std). (a)~AUROC and (b)~AUPRC for DeliMF, the structured-only model,
and two structured-input baselines. DeliMF's AUPRC gain concentrates on the
with-text and elderly subgroups, while it matches structured-only when no note
is present. The DeLLiriuM-style and E-PRE-DELIRIC-style baselines use only
structured inputs, so they are unaffected by note availability.}
\label{fig:subset}
\end{figure}

Splitting the test set by text availability sharpens the picture, as shown in
Fig.~\ref{fig:subset}. Among the patients who carry a note, slightly more than
half of the cohort, DeliMF attains an AUPRC significantly higher than the
structured-only model at $p=0.002$. Among those with no note, where the gate
closes and the model falls back on structure alone, it does not lose ground. The
framework thus adds signal where text exists and forfeits nothing where it does
not. Because nearly half the patients contribute no text, the whole-cohort AUROC
gain is modest by construction, which is why we rest the headline on AUPRC and on
the with-text population rather than on a sweeping AUROC claim.

\subsection{Age subgroup}
Stratifying the test set by age yields a clinically coherent pattern
(Fig.~\ref{fig:subset}). Delirium becomes both more common and harder to predict
with advancing age: incidence climbs from below ten percent in the younger group
to more than thirteen percent at age sixty-five and above, while AUROC drifts
downward across the same divide. DeliMF's gain settles where the burden is
heaviest. In the elderly subgroup its AUPRC reaches $0.366$ against $0.328$ for
the structured-only model, a gain of $3.8$ points, whereas below sixty-five the
two are essentially on par at $0.357$ versus $0.351$. The text
therefore helps most for the high-risk older population, which is precisely the
group a delirium early-warning tool is meant to serve.

\subsection{Interpretability}
\label{sec:interp}
\begin{figure}[t]
\centering
\includegraphics[width=0.78\columnwidth]{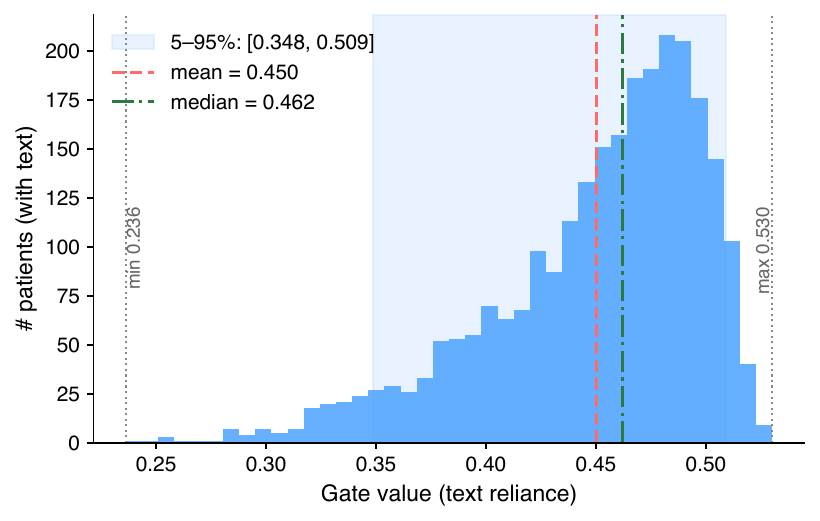}
\caption{Gate value distribution over patients with text. The mean of 0.45 and
the left tail show per-patient adaptive reliance on the text modality.}
\label{fig:gate}
\end{figure}

\textbf{Gate behavior.} Across patients with text, the gate centers slightly below
one half and spreads over a broad range, trailing off into a left tail of patients
the model reads mostly from structured signals (Fig.~\ref{fig:gate}). Rather than
settling on a fixed mixing constant, it behaves as an adaptive per-patient weight
on the text modality, which is exactly the behavior the design was meant to
produce.

\textbf{Case study.} The most revealing patients are those that the
structured-only model misses but DeliMF recovers. Across the test set, 16 truly
delirious patients receive a structured-only probability below 0.5 yet a DeliMF
probability above it. In every one of these cases, the note carries a recognized
delirium precipitant that the tabular record does not encode well: a head CT
reading ``subarachnoid hemorrhage'' with ``diffuse axonal injury''; a history of
``alcoholic cirrhosis'' under immunosuppression with suspected infection; ``acute
pneumonia'' in a hypoxic nonagenarian; a status-post ``subdural hematoma''
evacuation; or a report noting that the study was ``degraded by patient motion,''
with agitation itself being a harbinger of delirium. Representative examples are
listed in Table~\ref{tab:cases}. Collectively, these cases span the classic
precipitant categories of acute brain injury, infection, alcohol withdrawal, the
post-operative state, and advanced age. What sets these phrases apart is not the
diagnosis alone but its framing: an \emph{enlarging} hemorrhage, a consolidation
that \emph{could be} pneumonia, an infection flagged as a \emph{possible
precipitant}, or a scan \emph{degraded by motion}. Such dynamic and hedged
descriptions are precisely what a tabular feature cannot represent, and they
show concretely what the aggregate metrics report: the text supplies context the
numbers omit.

\begin{table}[t]
\caption{Representative cases the structured-only model misses but DeliMF catches
(true label: delirium). The last column quotes verbatim phrases from the first-24\,h
radiology report, showing that DeliMF recovers true delirium cases by extracting additional information.}
\label{tab:cases}
\centering
\footnotesize
\renewcommand{\arraystretch}{1.35}
\setlength{\tabcolsep}{4pt}
\begin{tabular}{>{\centering\arraybackslash}m{0.05\columnwidth} >{\centering\arraybackslash}m{0.09\columnwidth} >{\centering\arraybackslash}m{0.09\columnwidth} m{0.56\columnwidth}}
\toprule
\rowcolor{caseHdr}
\textcolor{white}{\textbf{Age}} & \textcolor{white}{\textbf{Struct.}} & \textcolor{white}{\textbf{DeliMF}} & \textcolor{white}{\textbf{Evidence in the radiology report}} \\
\midrule
\rowcolor{caseAlt}
61 & 0.21 & \best{0.56} &
\ev{1}{Parietal hemorrhage enlarging: ``$12\times12$\,mm'' vs.\ ``$8\times5.5$\,mm'' prior}
\ev{2}{``consistent with diffuse axonal injury''} \\
68 & 0.29 & \best{0.68} &
\ev{1}{``EtOH cirrhosis c/b ascites, varices,'' HCC s/p liver transplant}
\ev{2}{immunosuppressed ``on tacrolimus''}
\ev{3}{``possible infection as precipitant''} \\
\rowcolor{caseAlt}
91 & 0.37 & \best{0.69} &
\ev{1}{``Hypoxic respiratory failure''}
\ev{2}{right-lower-lobe consolidation that ``could be acute pneumonia''} \\
75 & 0.47 & \best{0.55} &
\ev{1}{S/p ``subdural hematoma evacuation,'' two burr holes}
\ev{2}{collection decreased: 19 vs.\ 26\,mm} \\
\rowcolor{caseAlt}
45 & 0.14 & \best{0.63} &
\ev{1}{C7 fracture with ``extensive edema''}
\ev{2}{study ``substantially degraded by patient motion'' (agitation)} \\
\bottomrule
\end{tabular}
\end{table}

%==============================================================
\section{Discussion}
%==============================================================
Our results indicate that clinical text helps where it exists, and that it helps
most on the metric reflecting positive-class quality. The whole-cohort AUROC gain
over the structured-only model is small and not statistically significant. This
outcome is expected rather than disappointing, since $44.6\%$ of patients
contribute no note and the available notes are radiology reports, an inherently
indirect signal. What the text does provide is a significant AUPRC gain on the
with-text population and, through the gate, robustness on the remainder. For a
deployment whose purpose is to surface high-risk patients for intervention,
precision-recall on the positive class is the quantity that matters most, and it
is precisely there that DeliMF improves on the structured-only model.

Several limitations bound these claims. First, radiology reports are only one
slice of the clinical narrative; integrating nursing notes or chief complaints,
for instance through concept-based text modeling~\cite{yang2025clmn}, is a
natural next step. Second, our evaluation is confined to the single-center
MIMIC-IV cohort and lacks external validation. Third, we do not test robustness
to erroneous or manipulated notes, to which language models are known to be
vulnerable~\cite{wang2025safety,wang2026clucert}. None of these limitations
undercuts the controlled comparisons within our fixed cohort and protocol.

%==============================================================
\section{Conclusion}
%==============================================================
DeliMF fuses structured ICU time series with real, leakage-safe clinical text
through a gated cross-attention module that adapts to each patient and degrades
gracefully when text is absent. Our experiments show that early clinical text
carries usable signal for ICU delirium prediction, that a learned per-patient
gate exploits this signal where it exists while protecting the model where it does
not, and that gated cross-attention extracts more from the text than naive fusion
or EHR serialization does. Together, these findings indicate that reading the
clinical notes is worth the effort, and they point toward richer note types and
external cohorts as promising next steps.

\bibliographystyle{IEEEtran}
\bibliography{refs}

\end{document}